\pdfoutput=1
\documentclass[11pt]{article}
\usepackage{coling2020}
\usepackage{times}
\usepackage{latexsym}

\usepackage{url}
\usepackage{color}

\usepackage{amsmath,amsfonts,amssymb}

\usepackage{tikz}
\usetikzlibrary{arrows,shapes,fit}
\usepackage{tabularx}
\usepackage{multirow}
\usepackage{makecell}
\newcolumntype{Y}{>{\centering\arraybackslash}X}
\newcolumntype{R}{>{\raggedleft\arraybackslash}r}
\newcolumntype{L}[1]{>{\raggedright\arraybackslash}p{#1}}
\newcolumntype{M}[1]{>{\raggedright\arraybackslash}p{#1}}
\newcolumntype{T}[1]{>{\raggedleft\arraybackslash}p{#1}}
\newcolumntype{P}[1]{>{\centering\arraybackslash}p{#1}}
\usepackage{siunitx}
\usepackage{todonotes}
\newcommand{\SideNote}[2]{} 
\renewcommand{\SideNote}[2]{\todo[color=#1,size=\small]{#2}}

\usepackage{enumitem}
\setlist[enumerate]{nosep, topsep=1pt}
\setlist[itemize]{nosep,topsep=1pt}
\definecolor{darkgreen}{rgb}{0.0, 0.5, 0.0}
\definecolor{Peach}{HTML}{F7965A}
\definecolor{CornflowerBlue}{HTML}{41B0E4}
\definecolor{OliveGreen}{HTML}{138a07}
\usepackage{booktabs}
\usepackage{hyperref}

\colingfinalcopy % Uncomment this line for the final submission

\title{Investigating Societal Biases in a Poetry Composition System}

\author{Emily Sheng\thanks{~~Work done while interning at Google} \\
  USC Information Sciences Institute \\
  Marina del Rey, CA \\
  {\tt ewsheng@isi.edu} \\\And
  David Uthus \\
  Google Research \\
  Mountain View, CA \\
  {\tt duthus@google.com} \\}

\date{}

\begin{document}
\setlength{\abovedisplayskip}{4pt}
\setlength{\abovedisplayshortskip}{4pt}
\setlength{\belowdisplayskip}{4pt}
\setlength{\belowdisplayshortskip}{4pt}

% The following footnote without marker is needed for the camera-ready
% version of the paper.
% Comment out the instructions (first text) and uncomment the 8 lines
% under "final paper" for your variant of English.
% 
\blfootnote{
    %
    % for review submission
    %
    % \hspace{-0.65cm}  % space normally used by the marker
    % Place licence statement here for the camera-ready version. See
    % Section~\ref{licence} of the instructions for preparing a
    % manuscript.
    %
    % % final paper: en-uk version 
    %
    % \hspace{-0.65cm}  % space normally used by the marker
    % This work is licensed under a Creative Commons 
    % Attribution 4.0 International Licence.
    % Licence details:
    % \url{http://creativecommons.org/licenses/by/4.0/}.
    % 
    % % final paper: en-us version 
    %
    \hspace{-0.65cm}  % space normally used by the marker
    This work is licensed under a Creative Commons 
    Attribution 4.0 International License.
    License details:
    \url{http://creativecommons.org/licenses/by/4.0/}.
}

\maketitle
\begin{abstract}
  There is a growing collection of work analyzing and mitigating societal biases in language understanding, generation, and retrieval tasks, though examining biases in creative tasks remains underexplored.
  Creative language applications are meant for direct interaction with users, so it is important to quantify and mitigate societal biases in these applications.
  We introduce a novel study on a pipeline to mitigate societal biases when retrieving next verse suggestions in a poetry composition system. 
  Our results suggest that data augmentation through sentiment style transfer has potential for mitigating societal biases.
\end{abstract}

\section{Introduction}
Our increasing reliance on natural language processing (NLP) tools to produce trustworthy and helpful information means we must also be increasingly vigilant to the social ramifications of NLP techniques.
Despite increasing attention to the ethical issues in NLP and development of techniques to mitigate biases in a variety of tasks, examining biases in creative NLP tasks remains underexplored; however, biases in creative tasks are equally as important.
The primary goal of creative NLP systems is to be disseminated in a society (e.g., for self expression and collective social enjoyment, education \cite{foster2008poetry}, therapy \cite{connolly2004healing}), whereas other NLP systems are primarily driven by some non-social goal (e.g., answer a query correctly or retrieve all relevant named entities).
Any existing societal biases propagated through creative systems have direct impact on our society. 
For example, biases in a system that is meant to educate about different perspectives may end up reinforcing demographic stereotypes.

In this work, we focus on quantifying and mitigating societal biases in a creative language application.
Specifically, we propose techniques to mitigate biases in the poetry composition system introduced by \newcite{uthus2019}. This system allows the user to collaboratively compose a poem using machine-suggested novel verses in the style of classic American poets.
As creative works are often shaped by the lived experiences and timely issues of the creator's life, a poetry composition system trained on poems from different authors of different eras may reflect a variety of societal biases.
Table~\ref{tab:examples} highlights the types of subtle differences in system responses when the user input contains different genders.

We define societal biases as unequal social perceptions of different socially-defined groups of people.
Propagating unequal representations of demographic groups reinforces representational harms, such as stereotypes, leading to discrimination and potential allocational harms, such as unequal job opportunities \cite{blodgett2020language,barocas2017problem}.\footnote{Allocational harms are when a system allocates resources unfairly to different groups.}
To evaluate societal biases, we examine the language polarity (i.e., sentiment) of the suggested verses when different demographic groups are mentioned in the user input.\footnote{We use sentiment as a proxy metric for the social perception of demographic groups.}
A system that favors different sentiment verse suggestions for mentions of different demographic groups (e.g., \textit{positive} for demographic \textit{A} and \textit{negative} for demographic \textit{B}) could propagate unequal positive and negative associations and amplify existing demographic inequalities.
However, even in the case where a model suggests verses with similar sentiment scores for different groups, the model can still propagate biases by reinforcing similar amounts of different negative stereotypes for each group.
Thus, we propose a technique to mitigate biases by making the verses suggested by the poetry system less negative in sentiment. 

In this preliminary study, we focus on retrieving less negative verses across different demographic groups to reduce harms from societal biases within and across demographic groups.\footnote{There could also be harms in other scenarios, e.g., if the user's input contains harmful content about a demographic and is followed by positive verse suggestions, or if demographic mentions occur in suggestions following negative verses, though we leave this to future work.}
Since there is no guarantee that verses with similar sentiment are biased or unbiased in the same way, we do not explicitly constrain equalizing the sentiments of verses suggested for different groups.
Results show that our method has promising results for both reducing negative verses and keeping the distribution of verse sentiments across groups comparable.

Our contributions are 1) a pipeline approach for mitigating societal biases in a poetry composition system and 2) a labeled poetry sentiment dataset.
For the first part of the pipeline, we introduce a poetry sentiment dataset and build a BERT-based \cite{devlin2019bert} sentiment analyzer for poetry. 
These sentiment tools are subsequently used to train a style transfer model \cite{li2018delete}, which is then used to augment data to train the next verse prediction component in the poetry composition system. 
Our results indicate that style transfer has potential as an augmentation technique to reduce societal biases.
Specifically, we can influence the model to suggest verses with more positive sentiment while keeping the suggested verse quality comparable.
This exploratory study introduces the capabilities of style transfer augmentation to mitigate biases and is an example of how bias mitigation can be applied to creative language tasks and information retrieval components.

\begin{table}[!t]
\small
\begin{center}
    \begin{tabular}{l l}
    \toprule
    \bf Suggested next verses for user input: \textit{The women} & \bf Suggested next verses for user input: \textit{The men} \\ \midrule
    
    \textit{\color{red}Hate to their love like that evil womanhood-} & \textit{\color{darkgreen}Looked the warm day in their manly way+} \\ \cmidrule(lr){1-2}
    \textit{\color{red}Drawed tears upon their wall-} &  \textit{\color{darkgreen}Lay on their crowns at their gracious whim+} \\ \cmidrule(lr){1-2}
    \textit{\color{red}Hide in the shame of their loveless men-} & \textit{Await, for the brand of their command} \\ \cmidrule(lr){1-2}
    \textit{Heard, if they were with flowers} & \textit{Brought to their manhood in the light-eyed} \\ \cmidrule(lr){1-2}
    \textit{Do with their wings and their dolls } & \textit{Ran with their rifles} \\ \bottomrule
    \end{tabular}
    \end{center}
    \vspace{-0.5em}
\caption{\label{tab:examples} Examples of verse suggestions for a user input of ``\textit{The women}'' or ``\textit{The men}'', where the suggestions for the former have more negative connotations (${\color{red}^-}$), and those for the latter have more positive connotations (${\color{darkgreen}^+}$). Even neutral examples contain gender stereotypes, though we focus on the negative and positive examples in this work.}
\vspace{-1em}
\end{table}
% - "The women"
% - Philip Freneau:"Spend as themselves as gay,", "Heard with their new influence men", "Be a slave of their ruin"
% - Oliver Wendell Holmes Sr.: "Pass on their royal tongues", "Stamp their haughty smile", "Stamp their scanty spear"
% - James Russell Lowell: "Master their young loins", "Wins from their snug", "Master their woman;"
% - "The men"
% - Philip Freneau: "Turned the freight from their lee,", "Slain very long than him their aims will", "Would have left at all their spirit hard"
% - Oliver Wendell Holmes Sr.: "Stamp their ancient Empire", "Gather their strong axe;", "Raised fast their strong hands"
% - James Russell Lowell: "Bear to their lot", "Feels while their beard", "Turned from their single"

\begin{figure}[t]
\centering
\scalebox{0.8}{
      \begin{tikzpicture}[
  font=\sffamily\footnotesize,
  every matrix/.style={ampersand replacement=\&,column sep=0.1cm,row sep=0.5cm},
  source/.style={draw,thick,fill=yellow!20,inner sep=.3cm},
  process/.style={draw,thick,rounded corners,inner sep=.3cm,fill=red!20},
  filter/.style={trapezium,draw,thick,trapezium angle=110,fill=blue!20},
  sink/.style={draw,cylinder,shape border rotate=90,aspect=0.2,fill=violet!20},
  dots/.style={gray,scale=2},
  to/.style={->,>=stealth',shorten >=1pt,semithick,font=\sffamily\footnotesize},
  every node/.style={align=center}]

  % Position the nodes using a matrix layout
  \matrix{
    \node[source] (input) {Input verse\\\textit{"The dogs"}}; \& \& \node[sink] (data) {Index of \\generated verses};  \\
    \node[process] (encode1a) {SentencePiece Tokenizer}; \& \& \node[process] (encode1b) {SentencePiece Tokenizer};  \\
    \node[process] (encode2a) {Transformer Layers}; \& \& \node[process] (encode2b) {Transformer Layers};  \\
    \node[process] (encode3a) {ReLu Layers}; \& \node[process] (search) {Dot Product}; \& \node[process] (encode3b) {ReLu Layers};  \\
    \& \node[source] (output) {Ranked next verses\\
    \textit{"Hold each vacant dog."}\\
    \textit{"Stir in their pictured chair"}\\
    \textit{"Suddenly with their claws"}\\
    \textit{"Float a single howl"}}; \& \\
  };

  % Draw the arrows between the nodes and label them.
  % \draw[to] (input) -- (rhyme);
  \draw[to] (input) -- (encode1a);
  \draw[to] (encode1a) -- (encode2a);
  \draw[to] (encode2a) -- (encode3a);
  \draw[to] (encode3a) -- (search);
  \draw[to] (data) -- (encode1b);
  \draw[to] (encode1b) -- (encode2b);
  \draw[to] (encode2b) -- (encode3b);
  \draw[to] (encode3b) -- (search);
  \draw[to] (search) -- (output);
  
  \tikzset{blue dotted/.style={draw=blue!50!white, line width=1pt,
                               dash pattern=on 1pt off 4pt on 6pt off 4pt,
                               inner sep=2mm, rectangle, rounded corners}};

  % Finally the blue dotted boxes are drawn as nodes fitted to other nodes
  \node (dotted box) [blue dotted, fit = (encode1a) (encode1b) (encode3b)] {};
  \node at (dotted box.north) [above, inner sep=3mm] {Next Verse Prediction};

  \node (dotted box) [blue dotted, fit = (data)] {};
  \node at (dotted box.north) [above, inner sep=3mm] {Verse Generation};
  
\end{tikzpicture}
    }
\vspace{-1em}
\caption{An overview of the poetry composition system, focusing on the components that make up the next verse prediction.}
\label{fig:overview}
\vspace{-1em}
\end{figure}
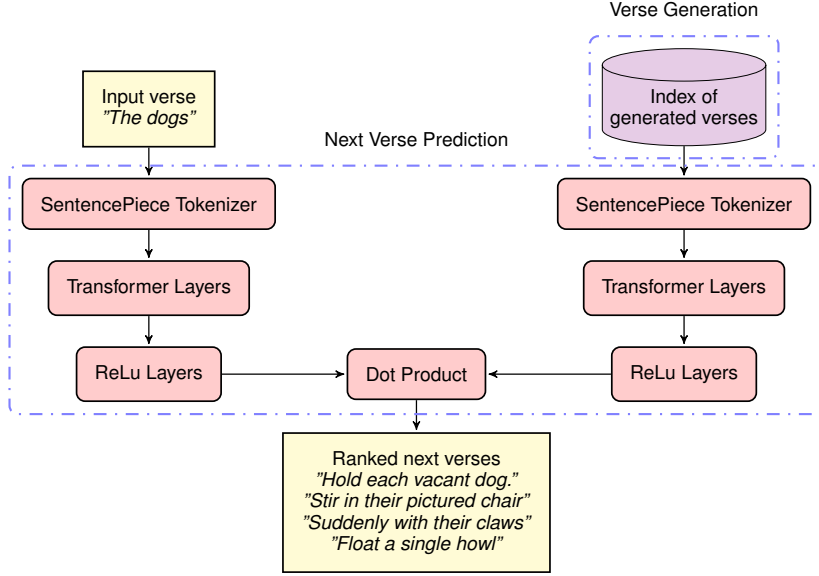

\section{Poetry Composition System}
\label{sec:poetry}
We investigate societal biases in the human-AI collaborative approach to composing poetry described by \newcite{uthus2019}.
In this setup, users compose a poem aided by suggestions from the poetry system.
Users can either directly use verse suggestions provided by the system, modify the suggestions, or create their own verses.
The suggested verses are generated in the style of various classic American poets (e.g., Walt Whitman, Emily Dickinson).

Figure~\ref{fig:overview} shows a schematic of the poetry composition system.
This system has two components:
\begin{itemize}
    \item \textbf{Verse Generation}: generates a large collection of verses in the styles of different poets and then indexes all verses for fast retrieval during poem composition.
    \item \textbf{Next Verse Prediction}: determines which pre-generated verses to suggest to the user, given a previous verse.
\end{itemize}

In a complex pipeline, there are multiple components to consider for the propagation of biases. 
For the poetry composition system, biases can propagate in both the language generation component and the next verse prediction component.
In this work, we only analyze biases in the next verse prediction component, because there is relatively little work examining biases in a retrieval task setting and because biases in the latter component could amplify biases from the earlier component.
Thus, beyond describing the verse generation component as a Transformer-based model trained on the same data as the next verse prediction dataset, we largely treat verse generation as a black box component.\footnote{The verses available to be retrieved from the verse generation component likely reflect distributional societal biases. As a first step towards making the set of available verses less gender-biased, we augment the original set of generated verses through counterfactual data augmentation \cite{lu2018gender}, i.e., swapping all female and male pronouns, and adding the resulting verses to the set of available verses.}
Note that the verse generation component is not free of grammatical and fluency issues, which are propagated down to the next verse prediction component.
For this work, we mainly focus on issues of societal biases and not grammatical issues stemming from the verse generation component.
In this section, we define components of the loss function and describe the training data for next verse prediction.

\paragraph{Next verse prediction loss.}
The next verse prediction model is a dual-encoder model similar to the one described by \newcite{henderson2017efficient}.
More specifically, let $\mathcal{R}$ be the entire fixed set of verse suggestions from the index of generated verses. Given a verse $x$, the model's goal is to search for the top $N$ responses $(y_1, y_2, ..., y_N) \in \mathcal{R}$, ordered by decreasing model probability
\begin{equation}
% \small
    \label{prob}
    P(y|x)=\frac{P(x,y)}{\sum_{r=1}^{\vert \mathcal{R} \vert} P(x, y_r)}.
\end{equation}
Eq.~\eqref{prob} requires summing over all possible responses $y_r$ when training, which is prohibitively expensive to calculate. Instead, we follow \newcite{henderson2017efficient} and sample $K$ responses to estimate $P(y|x)$ as
\begin{equation}
    \label{prob-est}
    % \small
    P_{approx}(y|x)=\frac{P(x,y)}{\sum_{k=1}^K P(x, y_k)}.
\end{equation}
In this work, the joint probability $P(x,y)$ is estimated with a learned scoring function $\mathcal{S}$, where $P(x,y)\propto e^{S(x,y)}$, so we can rewrite Eq.~\eqref{prob-est} as
\begin{equation}
    \label{score-prob-est}
    % \small
    P_{approx}(y|x)=\frac{e^{S(x,y)}}{\sum_{k=1}^K e^{S(x,y_k)}}.
\end{equation}
For the scoring function $\mathcal{S}$, we use a tower of Transformer \cite{vaswani2017attention} layers and feed-forward layers to encode $x$ and $y$ into the vectors $\mathrm{\bf{h}}_x$ and $\mathrm{\bf{h}}_y$, respectively.\footnote{Details and hyperparameters are in the Appendix.} The  dot product scoring function can then be written as $S(x,y)=\mathrm{\bf{h}}_x^\intercal \mathrm{\bf{h}}_y$.

Recall that to approximate the conditional probability $P_{approx}(y|x)$, we sample $K$ responses.
When training with a batch size of $K$, we treat $(x_i, y_i)$ as the positive example and $(x_i, y_k)$ where $k \neq i$ as negative examples, for $i,k \in [1, K]$.
From preliminary experiments, we find that including $(x_i, x_i)$ as a negative example improves retrieval, so we add this additional negative example for our experiments.
We define $\theta$ as the model's parameters.  
With this formulation, the average batch loss to minimize is the negative log probability 
\begin{equation}
    \label{loss}
    % \small
    \begin{aligned}
    \mathcal{L}(X,Y,\theta) &= -\frac{1}{K}\sum\limits_{i=1}^K \log P_{approx}(y_i \vert x_i) \\
    &= -\frac{1}{K}\sum\limits_{i=1}^K [S(x_i,y_i) - \log\sum\limits_{k=1}^K e^{S(x_i,y_k)}].
    \end{aligned}
\end{equation}

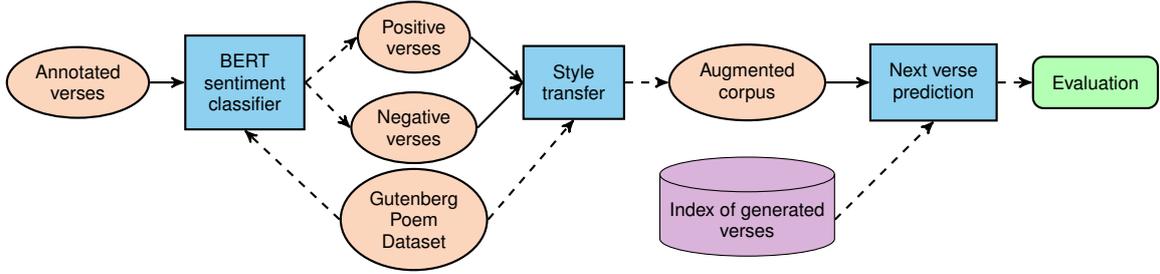
\begin{figure*}[t]
  \centering
  \begin{tikzpicture}[
  font=\sffamily\scriptsize,
  every matrix/.style={ampersand replacement=\&,column sep=0.45cm,row sep=-0.1cm},
  source/.style={draw,thick,fill=CornflowerBlue!60,inner sep=.25cm},
  data/.style={draw,ellipse,thick,fill=Peach!40,inner sep=.1cm},
  process/.style={draw,thick,rounded corners,inner sep=.25cm,fill=green!30},
  sink/.style={draw,cylinder,shape border rotate=90,aspect=0.2,fill=violet!30},
  to/.style={->,>=stealth',thick},
  tod/.style={->,dashed,>=stealth',thick},
  every node/.style={align=center}]

  % Position the nodes using a matrix layout
  \matrix{
    \node[data] (annotated) {Annotated\\verses}; \&
    \node[source] (bert) {BERT\\sentiment\\classifier}; \&
    \node[data,yshift=.6cm] (pos) {Positive\\verses}; \node[data,yshift=-.6cm] (neg) {Negative\\verses}; \&
    \node[source] (st) {Style\\transfer}; \&
    \node[data] (aug) {Augmented\\corpus}; \&
    \node[source] (nvp) {Next verse\\prediction}; \&
    \node[process] (eval) {Evaluation};  \\
    \& \& \node[data] (original) {Gutenberg\\Poem\\Dataset}; \& \& \node[sink] (index) {Index of generated\\verses}; \& \& \\
  };

  % Draw the arrows between the nodes and label them.
  \draw[to] (annotated) -- (bert);
  \draw[tod] (bert.east) -- (pos.west);
  \draw[tod] (bert.east) -- (neg.west);
  \draw[tod] (original.west) -- (bert.south);
  \draw[tod] (original.east) -- (st.south);
  \draw[to] (pos.east) -- (st.west);
  \draw[to] (neg.east) -- (st.west);
  \draw[tod] (st) -- (aug);
  \draw[tod] (index.east) -- (nvp.south);
  \draw[to] (aug) -- (nvp);
  \draw[tod] (nvp) -- (eval);
 
\end{tikzpicture}
  \caption{\textbf{A schematic of our technique for bias mitigation of the next verse prediction component in the poetry composition system.} Solid lines indicate training time; dashed lines indicate inference time.}
  \vspace{-1em}
  \label{fig:schematic}
%   \vspace{-1em}
\end{figure*}

\paragraph{Next verse prediction dataset.}
We collect a corpus of classic poems from Project Gutenberg\footnote{\url{https://www.gutenberg.org/}} to use as training data for the next verse prediction component --- we refer to this collected dataset as the Gutenberg Poem Dataset in this work and rely on the dataset for several components in our bias mitigation pipeline.
We split each poem into verses, and pair each verse with the subsequent verse to form the groundtruth data for next verse prediction.\footnote{Poems from Project Gutenberg are not stored in consistent formats, so we manually search the corpus for books of poems from famous classic American poets and then filter out malformatted poems, for a 45MiB dataset (more stats in Table~\ref{dataset-stats}).}

\section{Bias Mitigation Through Data Augmentation}
We propose a technique for bias mitigation through data augmentation of the training data.
Data augmentation has been used in previous works \cite{lu2018gender,zhao2018gender,park2018reducing} as a technique to mitigate societal biases in NLP systems. We use sentiment style transfer for data augmentation to target the retrieval of verses with less negative societal biases towards different demographic groups.
Our overall pipeline for augmentation and evaluation is shown in Figure \ref{fig:schematic}. By using style transfer for data augmentation instead of filtering out negative examples, we can circumvent data sparsity issues and promote model robustness. In the creative language domain, it can be difficult to obtain a large dataset, due to intellectual property and copyright restrictions. With style transfer-based augmentation, we can generate styled variations of our original data that can be more consistent in content and poet style.
In this section, we introduce definitions and then describe our proposed solution for bias mitigation through data augmentation.

\subsection{Definitions}
Demographic groups are socially-defined groups of people.
In the context of poem verses, we use demographic groups of people interchangeably with their mentions in text, e.g., the demographic group \textsc{gender-male} and the mention ``\textit{The man}''.
To more generally evaluate biases in suggested verses for different demographic groups, we curate a list of 25 demographic groups for various genders, race, and religions, e.g., ``\textit{The man}'', ``\textit{The African}'', ``\textit{The Muslim}''. 
Each demographic group is represented by two surface forms (one singular: ``\textit{The man}'' and one plural: ``\textit{The men}'').
We also create a list of 24 \emph{other} groups, where the goal is to obtain model suggestions for entities that are not a group of interest in the discussion of human societal biases. Specifically, we use a list of animals (e.g., ``\textit{The dog}'', ``\textit{The dogs}'').\footnote{\label{foot-appendix}More details in the Appendix.}

\begin{table*}[t]{
\small
\centering
 \begin{tabular}{L{4em} L{4em} L{4em} L{4em} L{20em}} \toprule
 \bfseries Sentiment score & \bfseries \# train samples & \bfseries \# dev samples & \bfseries \# test samples & \multirow{2}{*}{\bfseries Example} \\ \midrule
 
 negative & 155 & 19 & 19 & \textit{and that is why, the lonesome day,} \\ \cmidrule(lr){1-5}
 no impact & 555 & 69 & 69 & \textit{it flows so long as falls the rain,} \\ \cmidrule(lr){1-5}
 positive & 133 & 17 & 16 & \textit{with pale blue berries. in these peaceful shades--} \\ \bottomrule
 \end{tabular}
 \vspace{-0.5em}
 \caption{\label{dataset-stats} Dataset statistics and examples for different sentiment scores.}
}
% \vspace{-1em}
\end{table*}
% Full annotations kappa: 0.53
% 1550 samples
% ------------------
% Three annotations kappa: 0.58
% Three annotations Spearman’s correlation: 0.67
% 1369 samples
% ------------------
% 445 annotations for label: no-majority
% 693 annotations for label: 0
% 49 annotations for label: 2
% 166 annotations for label: 1
% 193 annotations for label: -1

\subsection{Sentiment Analysis}
In this section, we describe how we build a sentiment-labeled dataset of poem verses and use the dataset to train a sentiment classifier. This classifier is used both for generating training data for the style transfer model and the automatic evaluation of our bias mitigation technique.

\paragraph{Dataset.}
To the best of our knowledge, there is no existing public poetry dataset with sentiment annotations. 
We require an appropriate sentiment dataset in order to apply and evaluate our mitigation techniques, so we have two annotators label the sentiment of randomly picked verses from the Gutenberg Poem Dataset. The annotations use sentiment annotation guidelines described by \newcite{sheng2019woman}. For each sample, annotators could describe the language in the sample as \textit{negative}, \textit{no impact}, \textit{positive}, \textit{mixed (both negative and positive)}, or \textit{does not make sense}.$^{\ref{foot-appendix}}$

The inter-annotator agreement is 0.53 (Cohen's kappa) for 1550 annotated samples. If we remove samples where either annotator chose \textit{mixed} or \textit{does not make sense}, the kappa score increases to 0.58. Spearman's correlation for the samples with labels in the three sentiment categories (\textit{negative} = -1, \textit{no impact} = 0, \textit{positive} = 1) is 0.67. These correlations indicate decently strong inter-annotator agreement. 
For all annotated samples, we only keep the sample if there is agreement across both annotators and if the label is \textit{negative}, \textit{no impact}, or \textit{positive}.\footnote{The sentiment dataset can be found at \url{https://github.com/google-research-datasets/poem-sentiment}.}
Dataset statistics are in Table~\ref{dataset-stats}.

\paragraph{Sentiment classifier.}
We fine-tune a pretrained BERT model on the filtered annotated sentiment dataset. 
We use the uncased version of BERT base with a batch size of 32, learning rate of \num{1e-5}, maximum sequence length of 128, warmup proportion of 0.1, and train for 5 epochs.
The resulting sentiment classifier has a development set accuracy of 85.7\% and a test set accuracy of 84.6\%.
% - Hyperparams:  jellyfish TPU, 2x2 tpu topology, model-ckpt-80, 8 tpu cores

\subsection{Style Transfer}
With style transfer, we can automatically generate verses that are similar in content and poet writing style to existing verses yet different in sentiment style.
This section describes the style transfer technique and human evaluation of the technique.

\begin{table}[!t]{
\footnotesize
\centering
 \begin{tabular}{l l} \toprule
 \bfseries Input & \bfseries Generated Output \\ \midrule
 \textit{by those whose wrongs his soul had moved} & \textit{by those tender memory whose his own soul had moved} \\ \cmidrule(lr){1-2}
 \textit{the angel passed away} & \textit{the sweet singer angel passed away} \\ \cmidrule(lr){1-2}
 \textit{turnus 'tis true in this unequal strife} & \textit{turnus in this auspicious shore} \\ \cmidrule(lr){1-2}
 \textit{like warring giants angry huge and cruel} & \textit{like giants huge and sweet} \\ \cmidrule(lr){1-2}
 \textit{the darkness lingering oer the dawn of things} & \textit{lingering oer the dawn of many delicious things} \\ \bottomrule
 \end{tabular}
 \vspace{-0.5em}
 \caption{\label{style-examples-short}Examples of the inputs and generated positive sentiment outputs of the ``Delete, Retrieve, Generate'' style transfer technique \protect\cite{li2018delete} on poem verses.}
}
\vspace{-1em}
\end{table}

\paragraph{\textit{Delete, Retrieve, Generate} (DRG) technique.}
For style transfer, we follow the encoder-decoder model of \newcite{li2018delete}.
In this setup, the model deletes salient attribute markers (phrases that appear frequently in text of one style and not the other) in a text, retrieves an attribute marker of the opposite sentiment that appears in a similar context, and generates new text using the content of the original text and the new attribute marker.
The DRG model explicitly separates content and attributes of the text to facilitate the preservation of the words explicitly used in the original text, a characteristic especially desirable in the creative language domain where each word is carefully chosen to maximize creative expression. 
For example, for the verse ``\textit{like warring giants angry huge and cruel}'', deleting the negative attribute markers would result in ``\textit{like giants huge and}''. After retrieving positive attribute markers that are used in similar contexts, the model combines the original content and retrieved positive attribute markers to generate ``\textit{like giants huge and sweet}''. Examples of inputs and style-transferred outputs are in Table~\ref{style-examples-short}, and examples of different components of DRG are in the Appendix.

\paragraph{Experimental setup.}
By using the sentiment classifier to label verses in our Gutenberg Poem Dataset, we end up with 166K negative and 100K positive verses for the style transfer training set, and 19K negative and 11K positive verses for the evaluation set.
When training a DRG model with this dataset, the model learns to convert negative verses to positive verses and vice versa, though we only use negative to positive conversions for our data augmentation method.
Hyperparameter details are in the Appendix.

\paragraph{Human evaluation.} To evaluate the effectiveness of the DRG style transfer technique, we have humans manually annotate the 1) fluency of the style transferred text, and the 2) meaning preservation and 3) positive sentiment change between the original and new text.
All categories operate on a scale of 1 to 5.
For fluency, 1 means not fluent at all, and 5 means very fluent.
For meaning preservation, 1 means all original meaning is lost, and 5 means all meaning is preserved as much as possible, given the targeted sentiment change. 
For positive sentiment change, we ask how well the corresponding style transferred text became  more positive, with 1 meaning not more positive, and 5 meaning a lot more positive.
Since we are interested in making the suggested verses \emph{more positive}, we only measure the magnitude of how much more positive the style transferred text is compared to the original text.

\begin{figure}[!t]{
 \small
 \begin{center}
 \includegraphics[scale=0.37]{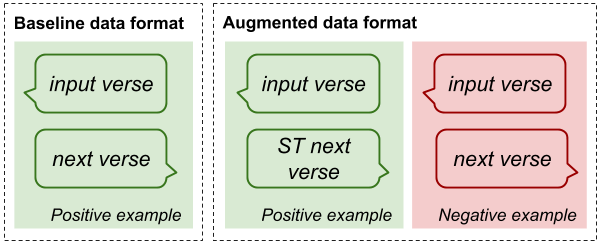}
 \end{center}
 }
 \vspace{-1em}
 \caption{\label{augmentation-example}
\textbf{Data augmentation details.} 
For example, \textit{input verse} = ``\textit{by the path an indian sat}'', \textit{next verse} = ``\textit{then i cried and ran away}'', and the positive sentiment style-transferred next verse \textit{ST next verse} = ``\textit{then i sing that human delight}''.
 (\textit{input verse}, \textit{next verse}) is the groundtruth data pair in the Gutenberg Poem Dataset. 
 In this example, \textit{input verse} contains a demographic mention (``\textit{indian}'') and \textit{next verse} has negative sentiment.
 The baseline next verse prediction model uses the original (\textit{input verse}, \textit{next verse}) pair as a positive example. The data augmentation model treats (\textit{input verse}, \textit{next verse}) as a negative example and uses (\textit{input verse}, \textit{ST next verse}) as the positive example.}
 \vspace{-1em}
 \end{figure}

Each of the 181 pairs of (original, style transferred) text is annotated by two annotators. Spearman's correlation is 0.73 for fluency, 0.85 for meaning preservation, and 0.82 for positive sentiment change. 
With these high inter-annotator correlation values, we average each sample score across the two annotators. 
The resulting average fluency of the style transferred text is 3.44, which means the new samples are moderately fluent. The average meaning preservation is 2.42, indicating that, on average, slightly more meaning is lost than preserved. 
The fact that the meaning of the verses are not as well-preserved through style transfer is expected, as poem verses tend to be difficult to interpret and are often expected to express multiple meanings. 
In the context of a creative language task, where inspiration and facilitating creativity is the main target, this weaker meaning preservation attribute is also more acceptable. 
Lastly, the average positive sentiment change is 2.22, which shows that the style transferred text is on average a bit more positive than the original text.

\begin{table}[!t]{
\footnotesize
\centering
 \begin{tabular}{l l l l l} \toprule
 \bfseries Data subset & \bfseries Next verse (-) & \bfseries Next verse (0) & \bfseries Next verse (+) & \bfseries Total \\ \midrule
 Input verse w/demo. & 2K (25\%) & 5K (63\%) & 1K (13\%) & 8K  (100\%) \\ \cmidrule(lr){1-5}
 Input verse w/o demo. & 41K (13\%) & 253K (79\%) & 25K (8\%) & 319K (100\%) \\ \bottomrule
 \end{tabular}
 \vspace{-0.5em}
 \caption{\label{next-verse-dataset}\textbf{Data statistics for the Gutenberg Poem Dataset used for next verse prediction (no augmentation)}. We divide the (input verse, next verse) pairs into those with an input that contains a demographic mention and those without. Within those groups, we show statistics for pairs with negative (-), neutral (0), and positive (+) sentiment next verses. Percentages are within each row.}
}
\vspace{-1em}
\end{table}

\subsection{Next Verse Prediction}
\label{ssec:next-verse-pred}

We use the trained style transfer model to augment the training data (from the Gutenberg Poem Dataset) for the next verse prediction model, and then run human and automatic evaluations to compare the original model with a model trained on the augmented data.

\paragraph{Data augmentation.}
We use the style transfer model to generate more positive sentiment verses from originally negative sentiment verses, and subsequently augment the training data with these generated verses.
More specifically, recall that the training data for the next verse prediction model is composed of \textit{input verse} and \textit{next verse} text pairs.
We conduct style transfer augmentation for specific training pairs:
\begin{enumerate}
    \item Pairs where the \textit{input verse} contains a demographic mention, e.g., ``\textit{by the path an indian sat}'', and the \textit{next verse} has negative sentiment, e.g., ``\textit{then i cried and ran away}''.
    \item Pairs where the \textit{input verse} does not contain a demographic mention and the \textit{next verse} has negative sentiment.
\end{enumerate}
We do not modify any training samples that are not in the above scenarios.
For the first scenario, we use the style transfer model to generate a positive version of the \textit{next verse}. 
The \textit{input verse} is then paired with the style-transferred positive \textit{next verse} as a positive example, and the original negative \textit{next verse} is used as a negative example for the \textit{input}.
We always generate a positive version of the \textit{next verse} if the \textit{input} contains a demographic mention, because that is our goal for bias mitigation.
For the second scenario, we randomly choose, with a probability of 0.5, whether to generate a positive \textit{next verse} and similarly augment.
With this random selection in the latter scenario, we end up generating new \textit{next verses} for approximately half of the samples; in subsequent results, we show that this amount of augmentation is enough to observe a difference in the results compared to the baseline model.\footnote{In preliminary results, style transfer augmentation on only the samples in the first scenario does not produce more positive sentiment verse suggestions overall. We hypothesize this lack of augmentation effectiveness is due to the relatively small amount of training pairs with a demographic mention in the \textit{input verse} (Table~\ref{next-verse-dataset}).}
Figure~\ref{augmentation-example} details how the augmented training data format differs from that of the baseline model.

Table~\ref{next-verse-dataset} shows statistics for the Gutenberg Poem Dataset.
Only about 2\% of all (\textit{input verse}, \textit{next verse}) pairs have \textit{inputs} that contain a demographic mention.
Furthermore, 25\% of the \textit{next verses} are negative for the pairs with a demographic in the \textit{inputs}, whereas only 13\% of \textit{next verses} are negative for the pairs without a demographic in the \textit{input}.
This imbalance highlights the opportunity for biases in the form of negative sentiment \textit{next verses} for \textit{inputs} containing demographic mentions.

\begin{table}[!t]{
\footnotesize
\begin{center}
 \begin{tabular}{L{2.7em} L{1.8em} L{2.2em} L{1.8em} L{2em} L{2em} L{2em}} \toprule
 \multirow{3}{*}{\bfseries \parbox{3em}{ \bfseries Prompt type}} & \multirow{3}{*}{\bfseries USE} & \multirow{3}{*}{\bfseries REL} & \multirow{3}{*}{\bfseries SEN} & \bfseries corr (SEN, USE) & \bfseries corr (SEN, REL)  &  \bfseries corr (REL, USE) \\ \midrule
 Min  & $+$1 & $-$4 & $+$1 & $-$1 & $-$1 & $-1$ \\ 
 Max & $+$5 & $+$4 & $+$5 & $+$1 & $+$1 & $+1$ \\ \cmidrule(lr){1-7}
 Demo. & 2.93 & $-$0.12 & 3.02 & 0.68 & 0.59 & 0.81 \\ 
 Others & 2.73 & $-$0.18 & 3.08 & 0.39 & 0.31 & 0.92 \\ \bottomrule
 \end{tabular}
 \end{center}
 }
 \vspace{-0.5em}
 \caption{\label{next-verse-results}\textbf{Human annotation results for next verse prediction.} For human comparisons of baseline and augmented model suggestions, given a demographic or other group mention in the input verse: USE = usability, REL = augmented model relevance $-$ baseline model relevance, SEN = sentiment. Spearman's correlation values (corr) are reported. Min is the lower-bound score; Max is the upper-bound score. USE/REL/SEN scores closer to Max indicate that the augmented model is better than the baseline.}
 \vspace{-1em}
 \end{table}

\paragraph{Experimental setup.}
The next verse prediction model and training procedure are as described in Sec.~\ref{sec:poetry}. Using the dual-encoder model with the loss in Eq.~\eqref{loss} and the augmented dataset, we train a style transfer-augmented next verse prediction model. Hyperparameters are in the Appendix.
For evaluation, we use ``\textit{The}'' followed by a demographic group or other group mention as the input verse (e.g., ``\textit{The man}'' or ``\textit{The dog}'') and perform human and automatic evaluations on the next verses suggested by the baseline and augmented versions of the next verse prediction component.

\paragraph{Human evaluation.}
To compare verses suggested by the style transfer-augmented model with verses suggested by the original model, we take the top 10 suggested verses from each model for each demographic mention and show them side-by-side to annotators.
Annotators are asked to label the relevance of both sets of suggestions, and compare the usability and sentiment between the two sets. Relevance is judged on a scale of 1 to 5, with 1 being not very relevant and 5 being very relevant. Usability is defined specifically for the task of composing a poem; 1 means the annotator is much more likely to use a suggestion from the original model, and 5 means the annotator is much more likely to use a suggestion from the augmented model.
For this creative language task, we include a measurement of usability as an alternate measurement of model performance.
The sentiment comparison score for this evaluation is also on a scale of 1 to 5, with 1 meaning the suggestions from the original model are more positive, and 5 meaning the suggestions from the augmented model are more positive.\footnote{Note that this is a different formulation from the style transfer evaluations.}
Two annotators annotate the relevance, usability, and sentiment comparison between the baseline and augmented model results for each of the 50 demographic mentions and 48 other group mentions. Spearman's correlation is 0.69 for usability, 0.66 for relevance, and 0.73 for sentiment change.

The averaged results over all demographic groups in Table~\ref{next-verse-results} suggest a slight increase in sentiment scores, which is the goal of our bias mitigation through augmentation.
We additionally see that the relevance and usability of the new samples are just slightly lower than those of the original samples, which could be due to the style transfer model's degree of accuracy in content preservation.
Measuring relevance and usability provides two qualitative dimensions of comparisons between the original and augmented data models.
However, we emphasize that text with high usability and relevance can still contain harmful stereotypes and societal biases, which is why sentiment is the main focus for our bias analysis.
On average, the style transfer augmentation is able to increase the sentiment of the retrieved verses with comparable usability scores across groups.

Table~\ref{next-verse-results} shows that relevance is highly but not perfectly correlated with usability and that sentiment is better correlated with the latter, indicating that using both relevance and usability may allow for more complete evaluations.
Usability as an evaluation metric can be more generally applicable to human-AI collaborative tasks.
For the task of selecting the best item to suggest to users, the metric of usability grounds evaluation to the specific values of the collaborative task at hand.

We also observe that the correlation between sentiment and usability for suggested verses for demographic prompts is nearly twice that of other prompts.\footnote{There is a similar pattern for the correlation between sentiment and relevance.}
This may indicate that annotators find that more positive sentiment suggestions are more helpful when the previous verse is about a demographic group --- perhaps annotators are able to better identify with humans than with non-human entities.
% Another possibility is a confounding factor where the annotators rate more positive verses as more usable for demographics because that is what they think the creators of the task want to see \cite{orne1962social}.

 \begin{table}[!t]{
 \small
 \begin{center}
 \begin{tabular}{L{3em} R R R R}
  \toprule
 \multirow{2}{*}{\bfseries \parbox{3em}{\centering \bfseries Prompt type}} & \multicolumn{2}{c}{\bfseries Baseline model} & \multicolumn{2}{c}{\bfseries Augmented model} \\ \cmidrule(lr){2-5}
 & \multirow{1}{*}{\bfseries Average} & \multirow{1}{*}{\bfseries Std. dev} & \multirow{1}{*}{\bfseries Average} & \multirow{1}{*}{\bfseries Std. dev} \\ \midrule
% all clumped together
\multirow{1}{*}{Demo.} & 0.01 & 0.12 &  0.06 & 0.13 \\ \cmidrule(lr){1-5}
\multirow{1}{*}{Other} & -0.01 & 0.16 & 0.01 & 0.12 \\ \bottomrule
%  \multirow{1}{*}{Demo-S} & 0.01 & 0.14 &  0.10 & 0.12 \\ \cmidrule(lr){1-5}
%  \multirow{1}{*}{Demo-P} & 0.00 & 0.10 & 0.01 & 0.13  \\ \cmidrule(lr){1-5}
%  \multirow{1}{*}{Other-S} & $-$0.03 & 0.16 & 0.02 & 0.14 \\ \cmidrule(lr){1-5} 
%  \multirow{1}{*}{Other-P} & 0.01 & 0.16 & 0.01 & 0.11 \\ \bottomrule
 \end{tabular}
 \end{center}
 }
 \vspace{-0.5em}
 \caption{\label{next-verse-results-auto}\textbf{Automatically labeled results for next verse prediction.} Sentiment average and standard deviations for suggested verses (sentiment scores $\in [-1, 1]$).
 The augmented model has higher average sentiment and lower or comparable standard deviations for both prompt types. This suggests the augmented model is effective at suggesting verses that are overall slightly more positive.
}
%  \vspace{-1em}
 \end{table}
 
 \begin{table*}[!t]{
\footnotesize
\centering
 \begin{tabularx}{\linewidth}{P{4.5em} L{19em} L{23em}} \toprule
 \bfseries Input & \bfseries Baseline model suggestions & \bfseries Augmented model suggestions \\ \hline
 \multirow{10}{*}{\textit{The woman}} & 
 \textit{\color{darkgreen}Uplifted than a manly me slain her strong,$^+$} & \textit{\color{red}This old man with toil, while her dog$^-$}\\
 & \textit{\color{darkgreen}Fate is life of her best she salt born.$^+$} & \textit{\color{darkgreen}Here from a hundred vain his blessing trust$^+$}\\
 & \textit{\color{red}She cannot the thirst dear, behind them back,$^-$} & \textit{Walking in the woody groves on earth,$^0$}\\
 & \textit{\color{darkgreen}Fate is life of him best Brought her scroll.$^+$} & \textit{Walking in the woody groves on earth.$^0$}\\
 & \textit{she wore the belt along the belt and tea.$^0$} & \textit{\color{darkgreen}she gives his prayer to every glorious thing,$^+$}\\
 & \textit{she wore the belt along the belt and tea;$^0$} & \textit{Olger to her End$^0$}\\
 & \textit{\color{darkgreen}Save the world when serious pine her strive.$^+$} & \textit{Reckon you have on every earth$^0$}\\
 & \textit{\color{darkgreen}Fate is life of her best that hath shed$^+$} & \textit{\color{darkgreen}Here from a hundred vain her blessing trust$^+$}\\
 & \textit{Go through thy small lessons long go this day,$^0$} & \textit{\color{darkgreen}House the fate their graceful own isle of work$^+$} \\
 & \textit{Whenever she goes back,$^0$} & \textit{\color{darkgreen}The gives the sunshine in her early sword!$^+$}\\ \cmidrule(lr){1-3}
 
 \multirow{10}{*}{\textit{The man}} &
 \textit{\color{red}Burns up every atom on his tongue$^-$} & \textit{Shall one alone him broad van pine and called and why,$^0$}\\
 & \textit{Great old groves send his cliffs$^0$} &  \textit{Shall one alone him broad van pine and called and why.$^0$}\\
 & \textit{Great in his labor been$^0$} & \textit{\color{darkgreen}Shall all the sacred summer thought and Creator,$^+$} \\
 & \textit{Whether an hour as kingdom spoke$^0$} & \textit{Shall one within so garden in his broad name$^0$}\\
 & \textit{States his steady nature took him son:$^0$} & \textit{She stood beside the brink, her brow they fell$^0$}\\
 & \textit{Great old groves send him cliffs$^0$} & \textit{Wrestles the soul$^0$}\\
 & \textit{States him steady nature took his son:$^0$} & \textit{Shall they who in him own road shall roam,$^0$}\\
 & \textit{\color{red}Burns up every atom on him tongue$^-$} & \textit{Due virtue hed an wants$^0$}\\
 & \textit{\color{red}Lest all the many is the haughty god,$^-$} & \textit{Life on his errand bound Fine woods$^0$}\\
 & \textit{\color{darkgreen}Shine over thy foe with lion forth$^+$} & \textit{Shall he who in him own road shall roam,$^0$}\\ \bottomrule
 \end{tabularx}
 \vspace{-0.5em}
 \caption{\label{verse-pred-examples}\textbf{Examples of user inputs and suggested verses from the baseline and augmented models.} Annotators labeled the overall sentiment change of the suggestions for ``\textit{The woman}'' as 4.0, and for ``\textit{The man}'' as 3.0 (on a scale of 1-5, 5 meaning the augmented model is much more positive). For more fine-grained automatic evaluation, {\color{darkgreen}$^+$} denotes positive, $^0$ denotes neutral, and {\color{red}$^-$} denotes negative sentiment. As expected, the human coarse-grained sentiment and automatic fine-grained sentiment scores are not perfectly correlated, yet both are useful (Sec.~\ref{ssec:next-verse-pred}).}
}
\vspace{-1em}
\end{table*}

\paragraph{Automatic evaluation.}
With human evaluations, we only show annotators a limited amount of verse suggestions for each demographic mention to not overwhelm the annotators.
% \cite{miller1956magical}.
However, it is also important to evaluate more retrieved verses, since users of the poetry composition system have the option to ask for more suggestions if the top retrieved ones are not inspiring enough.
% 5 if one poet is selected, 4 per poet if 2 poets selected, or 3 per poet if 3 poets selected.
With automatic evaluations, we can evaluate the top 50 suggested verses per demographic mention. The trade-off for this more comprehensive evaluation is that we rely on the automatic sentiment analyzer, which may not be as accurate as the human annotations. Also, the human evaluations directly compare the overall perceived sentiment change between the baseline and augmented model verse suggestion sets, whereas the automatic evaluations can be more fine-grained in providing a sentiment label for each verse.

Table~\ref{next-verse-results-auto} displays the average sentiment scores and standard deviations across demographic and other groups. 
The augmented model suggests verses with slightly higher sentiment on average, while all the standard deviations are comparable.
These results suggest that the augmented model is effective at suggesting verses that are overall slightly more positive than those of the baseline model.
% While we only augment samples with singular demographic mentions (``man'') in the next verse prediction training data, we also present evaluations for the plural demographic mentions (``men'') to better understand how our model generalizes.
% Overall, we see that the average sentiment of suggested verses increases and the standard deviation of average sentiment scores across demographics also decreases.
% The former observation suggests that the augmented model is effective at suggesting verses that are overall more positive than those of the baseline model.
% The latter observation about the decreasing standard deviation indicates that augmentation helps different demographic mentions achieve more equal suggestions, in terms of sentiment.
% For the plural demographic mentions (``Demo-P'' and ``Other-P''), it is not clear if the average sentiment of suggested verses has significantly increased from the baseline to the augmented model, and the standard deviation for ``Demo-P'' suggestions is not reduced by the augmented model.
% These results indicate that the augmentation technique perhaps does not generalize as well to unseen demographics.
Table~\ref{verse-pred-examples} provides a more comprehensive example of automatic fine-grained and human coarse-grained sentiment labels.
The fine-grained annotation indicates that suggestions for ``\textit{The woman}'' are roughly equal in sentiment across models, while suggestions for ``\textit{The man}'' are less negative for the augmented model.
In contrast, annotators labeled the overall sentiment change of the suggestions for ``\textit{The woman}'' as 4.0, and for ``\textit{The man}'' as 3.0.
Thus, the fine-grained and coarse-grained labels of sentiment scores are not perfectly correlated, but both are useful for a comprehensive evaluation. 
Future work could explore the reliability of human versus classifier judgments of sentiment and biases.

\section{Discussion}
In any complex pipeline with multiple components, downstream components can propagate and amplify errors and biases from upstream components.
The human and automatic evaluations show that the sentiment of the verses suggested by the augmented model are overall only slightly more positive than those suggested by the baseline model.
The small magnitude of this change could be due to the quality of our trained style transfer model or the amount of augmentation applied (not to all negative \textit{next verses}), among other factors.
The evaluation results highlight some of the challenges of mitigating biases in a pipeline system.

More generally, there are advantages and disadvantages to applying mitigation techniques at different points in a pipeline.
For example, we could also apply style transfer after the next verse component suggests a set of verses.
In doing so, there would be no need to re-train the next verse component.
The disadvantages are that we would have to style transfer all verse suggestions, thereby incurring more memory usage (apply style transfer beforehand and store the new verses) or more latency for the user (apply style transfer to retrieved verses lazily as the user requests suggestions).
By applying style transfer bias mitigation during training of the next verse prediction component, we can more efficiently add new verses to be retrieved by the component.

Our work provides a preliminary study of how style transfer techniques can be used to augment data in the context of a retrieval model.
We note that there can also be other types of societal biases in this poetry composition system, e.g., occupations that are biased towards specific genders, and other racial and gender stereotypes.
Future work includes looking into how style transfer could be used for these other biases.
For example, one could style transfer negative and positive stereotypes into more neutral associations, and subsequently use the latter associations as positive examples and the former stereotypes as negative examples for the next verse prediction model.

\section{Related Work}
\paragraph{Societal biases in NLP applications.}
Several recent works have analyzed and mitigated societal biases in word embeddings \cite{bolukbasi2016man,caliskan2017semantics,zhao2018learning,may2019measuring,gonen2019lipstick}. Others have also detailed how biases can occur in language understanding tasks, such as coreference resolution \cite{rudinger2018gender,zhao2018gender}, semantic role labeling \cite{zhao2017men}, abusive language detection \cite{park2018reducing}, sentiment analysis \cite{shen2018darling,kiritchenko2018examining}, and language modeling \cite{sheng2019woman,bordia2019identifying,pryzant2020automatically,huang2019reducing}.
To our knowledge, societal biases in creative applications has not been explored, despite the fact that creative application are often intended for social use (e.g., self expression, education).

\paragraph{Biases in information retrieval techniques.}
Earlier work on biases in information retrieval systems include examining biases in resume search engines \cite{chen2018investigating} and image search results \cite{kay2015unequal}.
More recently, researchers have proposed different metrics to measure the amount of group and individual fairness in ranking systems \cite{zehlike2017fa,biega2018equity,singh2018fairness,yang2020causal}. Additionally, \newcite{rekabsaz2020neural} compare how different neural ranking models differ in gender bias.
For future work, we could apply some of these previously proposed bias metrics that use weighting to account for the positions of the retrieved results.

\section{Conclusion}
We introduce an exploratory study on societal biases in a poetry composition application.
Although biases in creative language applications are underexplored, it is important to examine biases in these applications that are primarily intended for social use.
Our results indicate that style transfer has potential as an augmentation technique to reduce societal biases.
More broadly, our study introduces the capabilities of style transfer augmentation to mitigate biases and as an example of how bias mitigation can be applied to information retrieval components.

\section*{Acknowledgements}
The authors would like to thank Maria Voitovich, Noah Constant, and Vlad Imir for the very helpful discussions.
Additionally, the authors are grateful to Nanyun Peng, Kai-Wei Chang, Prem Natarajan, Jason Teoh, Mandy Guo, and anonymous reviewers for their feedback of this work.

\bibliographystyle{coling}
\bibliography{coling2020}

\clearpage
\appendix
\section{Appendix}
\subsection{Demographic and Other Groups}
\paragraph{Demographic groups}
\textit{white person, european, black person, african, american, indian, native, asian, caucasian, chinese person, hindu, hispanic person, indigenous person, hawaiian, islander, latino, latina, woman, man, girl, boy, christian, jewish person, muslim, buddhist}
\paragraph{Other groups}
\textit{dog, cat, horse, chicken, bear, bird, shark, snake, pig, lion, turkey, wolf, spider, rabbit, duck, deer, cow, monkey, lobster, ape, pony, eagle, dolphin, bison}

\subsection{General Annotation Details}
Two unique annotators completed the sentiment labeling task, and another two unique annotators completed the evaluation labeling tasks (for both style transfer and next verse prediction).
Future work includes expanding tasks to more annotators.

\subsection{Style Transfer}
\begin{table*}[!h]{
\footnotesize
\centering
 \begin{tabular}{L{8.5em} L{8.5em} L{8.5em} L{8.5em} L{8.5em}} \toprule
 
  {\multirow{1}{*}{\makecell{\shortstack{\bfseries Original Input}}}} & 
  {\multirow{1}{*}{\makecell{\shortstack{\bfseries Original Content}}}} & 
  {\multirow{1}{*}{\makecell{\shortstack{\bfseries Original Attributes}}}} & {\multirow{1}{*}{\makecell{\shortstack{\bfseries Retrieved Attributes}}}} & 
 {\multirow{1}{*}{\makecell{\shortstack{\bfseries Generated Outputs}}}} \\ \midrule
 by those whose wrongs his soul had moved & by those whose his soul had moved & wrongs & tender memory & by those tender memory whose his own soul had moved \\ \cmidrule(lr){1-5}
 the angel passed away & the angel passed away & -- & low sweet & the sweet singer angel passed away \\ \cmidrule(lr){1-5}
 turnus 'tis true in this unequal strife & turnus in this & 'tis true unequal strife & auspicious day & turnus in this auspicious shore \\ \cmidrule(lr){1-5}
 like warring giants angry huge and cruel & like giants huge and & warring angry cruel & -- & like giants huge and sweet \\ \cmidrule(lr){1-5}
 the darkness lingering oer the dawn of things & lingering oer the dawn of things & the darkness & many mansions & lingering oer the dawn of many delicious things \\ \bottomrule
 \end{tabular}
 \vspace{-0.5em}
 \caption{\label{style-examples}Examples of different components of the DRG style transfer technique on poem verses. In this technique, the original input is split into content and salient attribute text.}
}
\vspace{-1em}
\end{table*}

\paragraph{Model parameters.}
Unless stated otherwise, we use default parameters from the original DRG work by \newcite{li2018delete}.
With 1 NVIDIA Tesla V100 GPU, it takes a couple of days to train with our implementation of the style transfer model. We use a batch size of 256, a maximum sequence length of 30, a vocab size of 20K, and word-level tokenization. We train for 100,000 steps and evaluate for 100 steps.

For model-specific parameters, we use a word embedding dimension of 128, an attention mechanism, a bidirectional LSTM encoder with 1 layer of 512 hidden dimensions and a dropout of 0.2, an LSTM decoder also with 1 layer of 512 hidden dimensions, a beam search decoder with a beam width of 5 for the data augmentation Scenario \#1 in \ref{ssec:next-verse-pred} and a beam width of 3 for Scenario \#2. We also use a max norm of 3 for regularization and Adam for optimization with a learning rate of 0.0001.

For the parameters specific to the ``Delete, Retrieve, Generate'' architecture \cite{li2018delete}, we use $n$-gram attributes and an attribute salience threshold of 10.

\paragraph{Annotation guidelines.}
In this task, you will be evaluating how an original snippet of text was transformed into a new snippet of text by changing the sentiment of the text.
\begin{itemize}
    \item Fluency: for the new text, how do you rate the fluency, i.e., the quality and readability of the text, with 1 being not fluent at all and 5 being very fluent.
    \item Meaning Preservation: comparing the new text against the old text, and ignoring the change of style, how well does the new text preserve as much of the original meaning, with 1 being all meaning is lost and 5 being preserving as much as possible given the sentiment change?
    \item Sentiment Change: comparing the new text against the old text, how well did the sentiment of the new text become more positive, with 1 being not more positive and 5 being a lot more positive?
\end{itemize}

\subsection{Next Verse Prediction}
\paragraph{Model parameters.}
Both encoders use a SentencePiece subword tokenizer \cite{kudo2018} to tokenize input verses and then encode with a stack of Transformer \cite{vaswani2017attention} layers followed by a stack of feed-forward layers.
Each encoder stack consists of the following:
\begin{itemize}
    \item Transformer layers: 4 layers, each with 4 attention heads, and a hidden size of 1024
    \item Feed-forward layers: 2 layers (hidden size of 500), ReLu activation for first layer, SoftSign activation for final layer
\end{itemize}
Training consisted of 15,000,000 steps with a batch size of 100 and a learning rate of 0.01 for the first 10 million steps and 0.001 afterwards. In the Transformers layers, we used an attention dropout of 0.1 and ReLu dropout of 0.1.

\paragraph{Annotation guidelines.}
In this task, you will be evaluating poetic verse suggestions. You will be shown a possible line of verse, and two sets of possible candidates to follow that given line of verse.

\begin{itemize}
    \item Relevance: given the current verse, how relevant are the suggestions in group [A$\vert$B], with 1 being not relevant and 5 being very relevant.
    \item Sentiment: given the current verse, how much more positive is the sentiment of the suggestions on group B compared to group A, with 1 being A is much more positive and 5 being B is much more positive.
    \item Usability: given the current verse and assuming you were composing a poem, how much more likely would you use one the suggestions in group B compared to group A, with 1 being B is much less likely and 5 being B is much more likely.
\end{itemize}

\end{document}